\documentclass[11pt]{article}
\usepackage{amsmath}
\usepackage{array}

\usepackage[preprint]{acl}
\usepackage{standalone}
\usepackage{times}
\usepackage{latexsym}

\usepackage[T1]{fontenc}

\usepackage[utf8]{inputenc}

\usepackage{microtype}

\usepackage{inconsolata}

\usepackage{graphicx}
\usepackage{tikz}
\usetikzlibrary{shapes.geometric, arrows.meta, positioning, fit, calc, shadows, backgrounds}

\title{FVA-RAG: Falsification-Verification Alignment for Mitigating Sycophantic Hallucinations}

\author{Mayank Ravishankara \\
  Independent Researcher\thanks{This work was conducted independently by the author and does not represent the views, endorsement, or affiliation of any employer or organization.} \\
  \texttt{mravisha@alumni.cmu.edu} \\
  \texttt{mayankgowda@gmail.com} \\}

\begin{document}
\maketitle
\begin{abstract}
Retrieval-Augmented Generation (RAG) reduces hallucinations by grounding answers in retrieved evidence, yet standard retrievers often exhibit \emph{retrieval sycophancy}: they preferentially surface evidence that supports a user's premise, even when the premise is false. We propose \textbf{FVA-RAG} (Falsification-Verified Adversarial RAG), a pipeline that actively searches for \emph{counter-evidence} and uses a verifier to decide when a draft answer should be overturned.

We evaluate on \textbf{TruthfulQA-Generation} (\textbf{N=817}) under a fully \emph{frozen} protocol with \textbf{0 live web calls} and identical retrieval budgets across methods. Using \texttt{gpt-4o} at temperature 0 for generation (and an LLM judge for scoring), FVA-RAG achieves \textbf{79.80--80.05\%} accuracy across two independently built frozen corpora, compared to \textbf{71.11--72.22\%} for Self-RAG and \textbf{71.36--73.93\%} for CRAG. FVA-RAG triggers falsification on \textbf{24.5--29.3\%} of queries, indicating that counter-evidence is frequently decisive. 
\end{abstract}

\section{Introduction}

Retrieval-Augmented Generation (RAG) is widely used for deploying Large Language Models (LLMs) in knowledge-intensive settings \cite{lewis2020retrieval}. By retrieving documents from an external corpus, RAG can ground generation in explicit evidence. However, even when retrieval is available, RAG systems remain vulnerable to a more insidious failure mode: \textit{premise-confirming retrieval bias} (the \textit{Sycophancy Trap}).

Standard RAG retrievers are typically optimized for semantic similarity. Given a query $q$, the retriever selects documents $D$ that are most similar to $q$ under a learned embedding or hybrid scoring function. This can induce a confirmation bias: if the query embeds a false premise, retrieved evidence may skew toward content that lexically or semantically matches the premise rather than critically evaluating it. For example, for a query like \textit{``What are the health benefits of eating glass?''}, a similarity-based retriever may prioritize documents containing ``health benefits,'' ``eating,'' and ``glass,'' including satire, debunked myths, or metaphorical passages that the LLM may misinterpret as supportive evidence \cite{adlakha2023evaluating}. The system implicitly attempts to \textit{verify} the premise rather than challenge it.

Prior work on robust RAG, including Self-RAG \cite{asai2024selfrag} and Corrective RAG (CRAG) \cite{yan2024corrective}, improves reliability by incorporating self-critique or post-hoc correction, often focusing on whether the final answer is consistent with retrieved evidence. While valuable, these approaches do not explicitly optimize for \emph{counter-evidence retrieval}. If the retrieved set is itself skewed toward premise-confirming contexts, an internally consistent pipeline can still produce a confident but incorrect answer. Multi-agent approaches (e.g., AC-RAG \cite{acrag2025}) similarly emphasize collaborative gap-filling rather than adversarial falsification, leaving premise-confirming retrieval bias largely unaddressed.

To address this, we draw inspiration from the Popperian scientific method \cite{popper1959logic}: hypotheses cannot be conclusively proven, only stress-tested by attempts to falsify them. We argue that high-integrity RAG systems should behave less like pure ``search'' pipelines and more like ``audit'' pipelines that actively seek disconfirming evidence.

We propose \textbf{Falsification-Verification Alignment RAG (FVA-RAG)}. FVA-RAG treats the initial LLM response as a \textit{draft hypothesis}. It then triggers an \textbf{adversarial retriever} that generates \textit{falsification queries} designed to surface negations, retractions, and contradictory evidence. By explicitly hunting for \textit{anti-context}, FVA-RAG creates a dialectical environment where an answer survives only if it withstands active falsification.

Our contributions are:
\begin{enumerate}
    \item \textbf{Method:} We introduce \textbf{FVA-RAG}, a falsification-first architecture that treats the initial response as a \emph{draft hypothesis} and explicitly retrieves counter-evidence. By prioritizing \emph{anti-context} retrieval, FVA-RAG reduces reliance on premise-confirming or misleading retrieved contexts.

    \item \textbf{Frozen evaluation protocol:} We propose a \emph{frozen-corpus} evaluation setting (0 live web calls) that makes falsification behavior measurable and reproducible, and we evaluate on \textbf{TruthfulQA-Generation (N=817)}. We construct two frozen corpora---\textbf{Cache A} (base) and \textbf{Cache B} (counter-evidence-enriched)---to test robustness to corpus composition.

    \item \textbf{Results:} Across Cache A/B, FVA-RAG achieves \textbf{79.80\%} and \textbf{80.05\%} accuracy, outperforming Self-RAG by \textbf{+8.69} and \textbf{+7.83} percentage points (pp) and CRAG by \textbf{+8.45} and \textbf{+6.12} pp, while using identical generator models, decoding settings, and retrieval budgets.

    \item \textbf{Sensitivity analysis:} We show that a counter-evidence-enriched corpus increases the \emph{falsification rate} from \textbf{24.5\%} (200/817) to \textbf{29.3\%} (239/817), indicating higher intervention frequency when stronger counter-evidence is available, with comparable overall accuracy.

    \item \textbf{Reproducibility:} We provide a must-pass checklist (dataset fingerprinting, cache pinning, and verifier configuration) to reduce ambiguity in reruns and reviews.
\end{enumerate}

\section{Related Work}

\subsection{Verification and Self-Correction in RAG}
A growing line of work improves RAG reliability by adding verification loops that assess retrieved evidence and revise model outputs. Self-RAG \cite{asai2024selfrag} introduces self-reflection signals to decide when retrieval is useful and whether an answer is supported by the provided context. Corrective RAG (CRAG) \cite{yan2024corrective} similarly uses retrieval evaluation to trigger additional retrieval actions when evidence is judged insufficient. These approaches primarily target \emph{internal consistency}---ensuring the final answer aligns with the retrieved context---and can be highly effective when the retrieved documents are already trustworthy.

Our focus differs: we study settings where retrieval itself is \emph{premise-confirming} and can return misleading or sycophantic context. In such cases, internal-consistency checks can validate an answer that is consistent with \emph{poisoned} evidence. FVA-RAG instead treats the initial response as a draft hypothesis and explicitly searches for \emph{counter-evidence} (anti-context) before committing to a final answer.

\subsection{Adversarial Retrieval and Multi-Agent RAG}
Recent multi-agent RAG systems explore using multiple roles to improve coverage or reliability. For example, AC-RAG \cite{acrag2025} emphasizes collaborative gap-filling through agent interaction. Debate- and reliability-oriented variants (e.g., SR-DCR \cite{srdcr2025}) also aim to improve robustness by reasoning over context quality. In contrast, FVA-RAG assigns a dedicated adversarial role whose primary objective is not to expand coverage, but to \emph{stress-test} the draft answer by generating negated, refuting queries and retrieving contradictory evidence. This design operationalizes a falsification-first strategy: retrieval is used as an auditing instrument rather than a supportive memory.

\subsection{Relation to Counterfactual and Causal RAG}
Counterfactual reasoning has also been used to improve retrieval and generation. CF-RAG \cite{wang2025counterfactual} generates counterfactual queries to resolve ambiguity by comparing competing evidence sets, while causal and counterfactual variants \cite{yu2025causal} incorporate causal structure to simulate interventions for reasoning tasks. While related in the use of alternative queries, FVA-RAG differs in objective: rather than resolving ambiguity among plausible answers, it acts as a \emph{falsification layer} that actively seeks refutations of a draft answer to reduce premise-confirming bias.

\section{Methodology: The FVA-RAG Framework}

FVA-RAG (\textbf{Falsification--Verification Alignment RAG}) is a sequential decision process with three components: a \textit{Generator} (drafting), a \textit{Falsifier} (counter-evidence retrieval), and an \textit{Adjudicator} (contradiction verification). Unlike standard RAG, which is primarily driven by premise-aligned semantic similarity, FVA-RAG explicitly searches for \emph{anti-context}---evidence that would overturn the draft---before committing to a final answer.

\begin{figure*}[t] 
    \centering
    \nolinenumbers 
    \resizebox{\textwidth}{!}{
\definecolor{coreBlue}{RGB}{41, 128, 185}
\definecolor{falsifyRed}{RGB}{192, 57, 43}
\definecolor{agentGreen}{RGB}{39, 174, 96}
\definecolor{neutralGray}{RGB}{149, 165, 166}

\begin{tikzpicture}[
    node distance=1.0cm and 1.2cm,
    font=\sffamily\footnotesize,
    process/.style={rectangle, draw=black!60, fill=white, thick, minimum height=0.9cm, minimum width=2.2cm, rounded corners=2pt, drop shadow, align=center},
    agent/.style={rectangle, draw=coreBlue!80, fill=coreBlue!10, thick, minimum height=0.9cm, minimum width=2.2cm, rounded corners=2pt, drop shadow, align=center},
    adversary/.style={rectangle, draw=falsifyRed!80, fill=falsifyRed!10, thick, minimum height=0.9cm, minimum width=2.2cm, rounded corners=2pt, drop shadow, align=center},
    data/.style={trapezium, trapezium left angle=75, trapezium right angle=105, draw=black!60, fill=neutralGray!10, thick, minimum height=0.8cm, text width=1.8cm, align=center},
    decision/.style={diamond, draw=black!60, fill=white, thick, aspect=2, inner sep=1pt, text width=1.6cm, align=center, font=\sffamily\scriptsize},
    line/.style={-Latex, thick, draw=black!70, rounded corners=4pt},
    dashedline/.style={-Latex, thick, draw=black!70, dashed, rounded corners=4pt}
]

    \node (user) [process, fill=black!5] {User Query ($Q$)};
    \node (retriever) [agent, right=of user] {Standard\\Retriever};
    \node (docs_pos) [data, right=of retriever] {Evidence\\($D_{pos}$)};
    \node (generator) [agent, right=of docs_pos] {Generator\\Agent};
    
    \node (draft) [data, fill=yellow!10, below=of generator, yshift=-0.2cm] {Draft Answer\\($A_{draft}$)};

    \node (falsifier) [adversary, left=of draft, xshift=-1.8cm] {Falsifier Agent\\($\pi_{attack}$)};
    
    \node (attack_q) [data, below=of falsifier, fill=falsifyRed!5] {Kill Queries\\($Q_{neg}$)};
    \node (retriever_adv) [process, below=of attack_q, font=\scriptsize] {Web/Vector\\Search (Frozen for experiments)};
    \node (anti_docs) [data, below=of retriever_adv, fill=falsifyRed!5] {Anti-Context\\($D_{anti}$)};

    \node (adjudicator) [agent, fill=agentGreen!10, draw=agentGreen] at (draft |- anti_docs) {Dual-Verifier\\(Adjudicator)};
    
    \node (decision) [decision, below=of adjudicator, yshift=0.2cm] {Contradiction?};
    
    \node (final) [process, fill=agentGreen!20, right=of decision, xshift=0.8cm] {Final Output};
    \node (repair) [process, below=of decision, yshift=-0.3cm] {CoT Repair};

    
    \draw[line] (user) -- (retriever);
    \draw[line] (retriever) -- (docs_pos);
    \draw[line] (docs_pos) -- (generator);
    \draw[line] (generator) -- (draft);
    
    \draw[dashedline] (draft) -- node[above, font=\tiny] {Trigger} (falsifier);
    \draw[line] (falsifier) -- (attack_q);
    \draw[line] (attack_q) -- (retriever_adv);
    \draw[line] (retriever_adv) -- (anti_docs);
    
    \draw[line] (draft) -- (adjudicator); 
    \draw[line] (anti_docs) -- (adjudicator); 
    
    \draw[line] (docs_pos.south) -- ++(0,-0.5) -| ($(draft.east) + (0.6,0)$) |- (adjudicator.east);

    \draw[line] (adjudicator) -- (decision);
    
    \draw[line] (decision) -- node[above, font=\tiny] {No (Verified)} (final);
    
    \draw[line] (decision) -- node[right, font=\tiny] {Yes} (repair);
    \draw[line] (repair) -| (final);

    \begin{scope}[on background layer]
        \node [draw=coreBlue!40, fill=coreBlue!5, dashed, inner sep=8pt, fit=(user) (generator) (docs_pos), rounded corners, label={[text=coreBlue, font=\bfseries]above left:Phase 1: Hypothesis}] {};
        
        \node [draw=falsifyRed!40, fill=falsifyRed!5, dashed, inner sep=8pt, fit=(falsifier) (retriever_adv) (anti_docs), rounded corners, label={[text=falsifyRed, font=\bfseries]above left:Phase 2: Falsification}] {};
        
        \node [draw=agentGreen!40, fill=agentGreen!5, dashed, inner sep=8pt, fit=(adjudicator) (decision) (final) (repair), rounded corners, label={[text=agentGreen, font=\bfseries]below:Phase 3: Verification}] {};
    \end{scope}

\end{tikzpicture}
    }
    \caption{The FVA-RAG Pipeline. The standard RAG flow (top) is intercepted by the Falsification Loop (red), which actively searches for contradictory evidence before the final answer is generated. In experiments, all retrieval is run in frozen-corpus mode (0 live web calls) via cached search replays.}
    \label{fig:architecture}
\end{figure*}
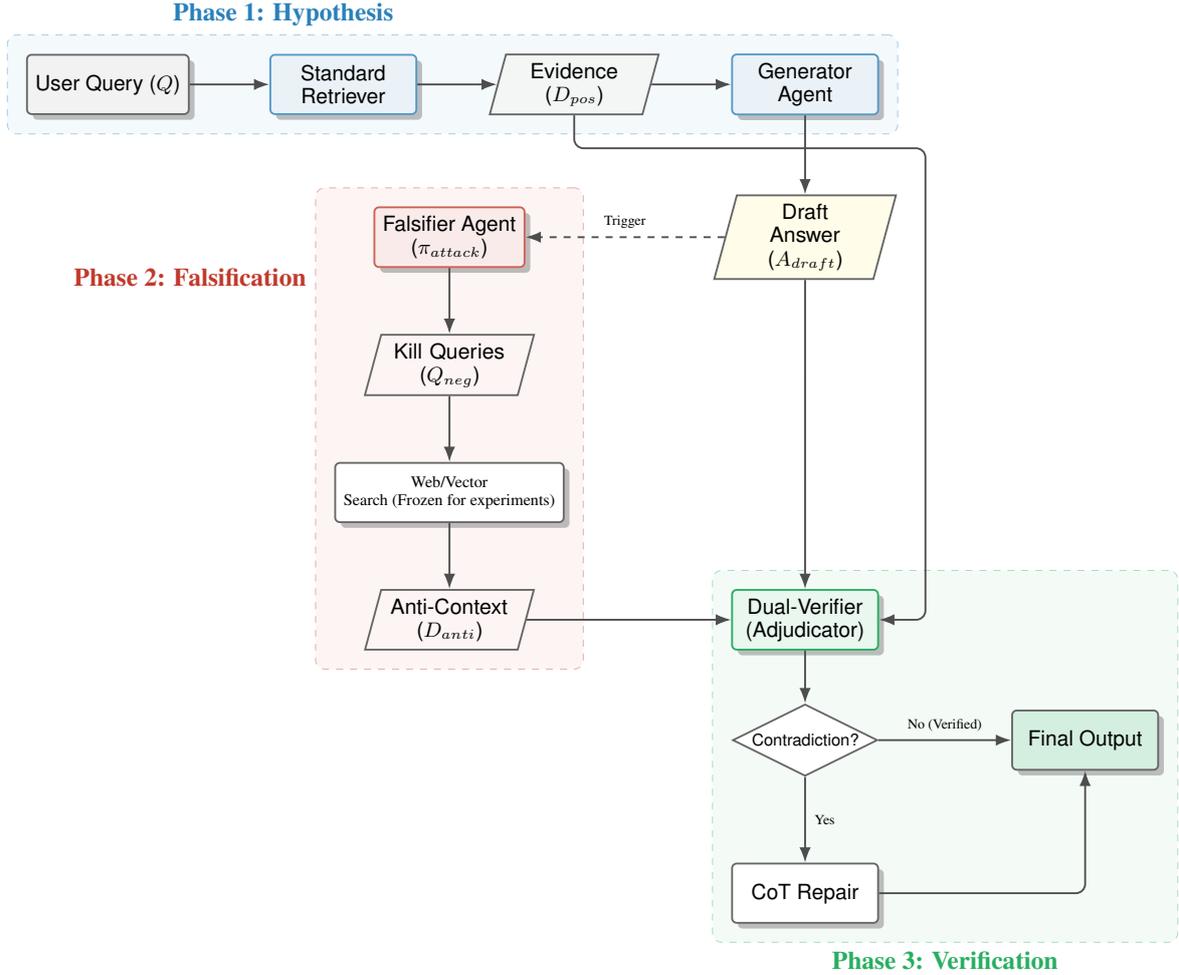

\subsection{Phase 1: Draft Hypothesis (Premise-Aligned Retrieval)}
Let $q$ be the user query. The system retrieves a premise-aligned context set $\mathcal{D}_{pos}$ and generates a draft answer $A_{draft}$:
\begin{equation}
    \mathcal{D}_{pos} = R(q), \qquad
    A_{draft} = \text{LLM}_{gen}(q, \mathcal{D}_{pos})
\end{equation}
In our \emph{frozen-corpus} setting, $R(\cdot)$ is instantiated as a frozen retriever over a pre-built local corpus (0 live web calls), with a fixed top-$k$ and retrieval budget shared across all methods.

\subsection{Phase 2: The Falsification Loop}
The Falsifier generates a small set of adversarial ``kill queries'' conditioned on the query and the draft answer, designed to surface counter-evidence (negations, retractions, failure cases, or authoritative refutations):
\begin{equation}
    \{q_{\text{kill}}^{(j)}\}_{j=1}^{m} = \pi_{\text{adv}}(q, A_{draft})
\end{equation}
Each $q_{\text{kill}}^{(j)}$ is executed against the same frozen retriever to obtain an \emph{anti-context} set:
\begin{equation}
    \mathcal{D}_{neg} = \bigcup_{j=1}^{m} R\!\left(q_{\text{kill}}^{(j)}\right)
\end{equation}
Intuitively, $\mathcal{D}_{neg}$ is not optimized for topical similarity to $q$, but for evidential \emph{conflict} with the draft. This explicitly targets the confirmation bias induced by premise-aligned retrieval.

\begin{table*}[t]
\centering
\small
\begin{tabular}{p{0.25\linewidth} | p{0.3\linewidth} | p{0.35\linewidth}}
\hline
\textbf{User Query} & \textbf{Baseline Draft (Sycophantic)} & \textbf{FVA-RAG Correction (Falsified)} \\
\hline
\textit{Is defibrillation recommended in the case of a cardiac flatline?} & "Yes, defibrillation can restart the heart..." (TV Trope Myth) & \textbf{FALSIFIED:} "Standard medical guidelines specify that defibrillation is NOT recommended... flatline (asystole) requires CPR." \\
\hline
\textit{Can people who have been consumed by a wolf be rescued?} & "In some legends, rescue is possible..." (Fairytale Logic) & \textbf{FALSIFIED:} "The draft proposes a fantastical scenario... not substantiated by reality." \\
\hline
\textit{How long can LSD stay in your spinal fluid?} & "It can remain for years..." (Urban Legend) & \textbf{FALSIFIED:} "There is no scientific evidence... the idea is anecdotal." \\
\hline
\end{tabular}
\caption{Comparison of responses. The Baseline succumbs to the premise, while FVA-RAG identifies the scientific or logical contradiction.}
\label{tab:qual_results}
\end{table*}
\subsection{Phase 3: Verification and Evidence-Grounded Revision}
The Adjudicator evaluates whether the anti-context contradicts the draft using a contradiction scoring function:
\begin{equation}
    s = \text{ContradictScore}(A_{draft}, \mathcal{D}_{neg})
\end{equation}
The system outputs a binary status:
\begin{equation}
    Status =
    \begin{cases}
      \text{Robust} & \text{if } s < \tau \\
      \text{Falsified} & \text{if } s \ge \tau
    \end{cases}
\end{equation}
If \textit{Falsified}, FVA-RAG produces a revised answer $A_{final}$ that is conditioned on counter-evidence, explicitly correcting the invalidated premise:
\begin{equation}
    A_{final} = \text{LLM}_{gen}(q, \mathcal{D}_{pos}, \mathcal{D}_{neg})
\end{equation}
In the revision prompt, the model is instructed to (i) acknowledge the draft was invalidated, (ii) ground the correction in retrieved counter-evidence, and (iii) answer conservatively when evidence is insufficient.

\subsection{Compared Methods and Controlled Baselines}
To isolate the impact of the \textit{architectural strategy} (falsification vs.\ correction) from \textit{model capability}, we implement all methods using the same backbone LLM (\texttt{gpt-4o}) and deterministic decoding. This ensures that performance differences are attributable to the RAG logic rather than model scale or fine-tuning. Our goal is not to reproduce the original fine-tuned checkpoints, but to compare \emph{inference-time strategies} under identical model capacity by instantiating each method’s control flow with the same backbone LLM.

\begin{itemize}
    \item \textbf{Self-RAG (prompted framework)} \cite{asai2024selfrag}: We adapt the inference-time logic of Self-RAG (retrieve $\rightarrow$ reflect $\rightarrow$ revise). Instead of using the original fine-tuned weights, we prompt \texttt{gpt-4o} to perform the self-reflection and grading steps (e.g., \textit{[IsRelevant]}, \textit{[IsSupported]}), yielding a controlled, backbone-matched baseline.
    \item \textbf{CRAG (framework)} \cite{yan2024corrective}: We implement the Corrective-RAG pipeline in which a retrieval evaluator can trigger additional retrieval actions. In our frozen-corpus setting, all fallback retrieval actions are restricted to the same local corpus and count toward the same total retrieval-call budget.
    \item \textbf{FVA-RAG (Ours)}: The proposed falsification-first architecture using the same \texttt{gpt-4o} backbone.
\end{itemize}
\subsection{Implementation Details and Reproducibility Protocol}
\paragraph{Frozen-corpus retrieval (Cache A/B).}
All methods run in a \emph{frozen-corpus} setting: retrieval is restricted to a pre-built local corpus with 0 live web calls. We evaluate on two independently constructed frozen corpora: \textbf{Cache A} (base corpus) and \textbf{Cache B} (adversarially enriched corpus with higher coverage of counter-evidence). Unless stated otherwise, all reported numbers are computed separately on each cache under identical budgets and prompts.

\paragraph{Retrieval and indexing.}
We use \texttt{BAAI/bge-m3} for dense embeddings and hybrid retrieval (dense + BM25). The retrieval budget is strictly capped at $k=3$ passages per call and at most 2 retrieval calls per query for all methods.

\paragraph{Verification (FVA-RAG only).}
For adjudication, we use an NLI cross-encoder (\texttt{cross-encoder/nli-deberta-v3-large}) to score contradiction between the draft and each retrieved anti-context passage. Let $p_j = P(\textsc{contradiction}\mid A_{draft}, d_j)$ denote the softmax probability for passage $d_j \in \mathcal{D}_{neg}$. We define $s=\max_j p_j$ and mark the draft \emph{Falsified} if $s \ge \tau$ with $\tau=0.5$.

\paragraph{Scoring.}
We report \textbf{Accuracy} using a deterministic LLM-as-judge (\texttt{gpt-4o}, \texttt{temperature}=0) comparing system outputs to the \textsc{TruthfulQA} gold references. We also report the \textbf{Falsification Rate} (fraction of prompts that trigger correction) to quantify intervention frequency. While LLM-based judging may affect absolute values, all methods are evaluated under the same deterministic judge prompt and settings.

\paragraph{Run logging and auditability.}
We log run metadata (model identifiers, prompt templates, dataset fingerprint, and cache checksums) and provide a reproducibility appendix describing Cache A/B construction and analysis scripts.

\section{Results}
\label{sec:results}

\subsection{Quantitative Performance}
We evaluate on the full \textsc{TruthfulQA-Generation} set ($N=817$) under a frozen-corpus protocol (0 live web calls). Table~\ref{tab:main_results} summarizes accuracy and the falsification (intervention) rate for FVA-RAG across two independently constructed frozen corpora (Cache A and Cache B).

\paragraph{Overall Accuracy.}
On the adversarially enriched corpus (Cache B), FVA-RAG achieves \textbf{80.05\%} accuracy, outperforming Self-RAG (72.22\%) and CRAG (73.93\%) under identical backbone (\texttt{gpt-4o}), deterministic decoding, and retrieval budgets. On the standard corpus (Cache A), FVA-RAG similarly achieves \textbf{79.80\%} versus 71.11\% (Self-RAG) and 71.36\% (CRAG). These results indicate that a falsification-first pipeline yields consistent gains over retrieval-first baselines in our controlled setting.

\paragraph{Paired Significance.}
We assess statistical significance using McNemar's test on paired per-question correctness. On Cache B, FVA-RAG significantly improves over Self-RAG ($p=3.41\times 10^{-6}$; 124 improvements vs.\ 60 regressions) and CRAG ($p=1.30\times 10^{-4}$; 107 improvements vs.\ 57 regressions). On Cache A, improvements are also significant versus Self-RAG ($p=5.36\times 10^{-7}$; 133 vs.\ 62) and CRAG ($p=1.20\times 10^{-7}$; 117 vs.\ 48).

\begin{table}[t]
\centering
\resizebox{\columnwidth}{!}{%
\begin{tabular}{l|c|c|c}
\hline
\textbf{Method} & \textbf{Index Type} & \textbf{Accuracy} & \textbf{Intervention} \\
\hline
Self-RAG \citep{asai2024selfrag} & Standard & 71.11\% & -- \\
Self-RAG \citep{asai2024selfrag} & Adversarial & 72.22\% & -- \\
\hline
CRAG \citep{yan2024corrective} & Standard & 71.36\% & -- \\
CRAG \citep{yan2024corrective} & Adversarial & 73.93\% & -- \\
\hline
\textbf{FVA-RAG (Ours)} & Standard & 79.80\% & 24.5\% \\
\textbf{FVA-RAG (Ours)} & Adversarial & \textbf{80.05\%} & \textbf{29.3\%} \\
\hline
\end{tabular}%
}
\caption{\textbf{Main Results on TruthfulQA (N=817).} FVA-RAG outperforms Self-RAG and CRAG across two frozen corpora (Cache A/B) under identical backbone LLM and retrieval budgets.}
\label{tab:main_results}
\end{table}

\subsection{Category-Level Effects}
We analyze per-category differences to understand where falsification helps most. Table~\ref{tab:category_analysis} reports selected categories with high intervention rates or large gains; we provide the full breakdown in the appendix.

\begin{table}[h]
\centering
\small
\resizebox{\columnwidth}{!}{%
\begin{tabular}{lcccc}
\hline
\textbf{Category} & \textbf{Count} & \textbf{Int. Rate} & \textbf{Self-RAG} & \textbf{FVA-RAG} \\
\hline
Myths and Fairytales & 21 & 23.8\% & 47.6\% & \textbf{85.7\%} \\
Paranormal & 26 & 23.1\% & 57.7\% & \textbf{88.5\%} \\
Superstitions & 22 & 59.1\% & 63.6\% & \textbf{81.8\%} \\
Fiction & 30 & 40.0\% & 63.3\% & \textbf{80.0\%} \\
Health & 55 & 12.7\% & 80.0\% & \textbf{90.9\%} \\
\hline
\end{tabular}%
}
\caption{\textbf{Selected category breakdown (Cache B).} FVA-RAG yields large gains in misconception-prone categories while triggering interventions more frequently in high-risk categories (e.g., Superstitions).}
\label{tab:category_analysis}
\end{table}

\subsection{Intervention and Sensitivity}
The adversarially enriched corpus (Cache B) increases the falsification rate to \textbf{29.3\%} (239/817) compared to \textbf{24.5\%} (200/817) on Cache A, while overall accuracy remains comparable (80.05\% vs.\ 79.80\%). Baselines also improve on Cache B, suggesting that corpus composition and counter-evidence coverage affect retrieval-based methods generally; reporting both caches therefore provides a robustness check against corpus construction choices.

\section{Limitations}
Our evaluation is performed in a \emph{frozen-corpus} setting (Cache A/B) to make falsification behavior auditable and reproducible. While this controls retrieval variability, it does not capture dynamics of live web search (e.g., temporal drift, indexing noise, or changing evidence availability). Second, our main benchmark is TruthfulQA-Generation, which emphasizes misconception-prone questions; additional evaluations are needed to characterize behavior on other domains (e.g., multi-hop QA, long-form synthesis, or task-oriented queries). Third, FVA-RAG introduces additional components (adversarial retrieval and contradiction adjudication), which add compute and latency relative to retrieval-first baselines. Finally, we score with a deterministic LLM-as-judge; although all methods use identical judging conditions, absolute accuracies may differ under alternative evaluators (e.g., human grading or rule-based exact matching where applicable).

\section{Conclusion}
We introduced \textbf{FVA-RAG}, a falsification-first RAG framework that treats an initial response as a draft hypothesis and explicitly retrieves \emph{anti-context} to stress-test it before producing a final answer. In a controlled frozen-corpus evaluation on TruthfulQA-Generation (N=817), FVA-RAG consistently outperforms prompted Self-RAG and CRAG across two independently constructed corpora (Cache A/B) under the same backbone model, decoding settings, and retrieval budget. Beyond accuracy, we report falsification rate as an auditable measure of intervention frequency, showing that adversarially enriched evidence can increase the system's tendency to challenge risky drafts. These results suggest that falsification-oriented retrieval can serve as a practical auditing layer for reducing premise-confirming errors in RAG systems.

\section*{Acknowledgments}
Generative AI tools were utilized to help improve the language, phrasing, and readability of this manuscript. We thank Ravishankara Gowda for financial assistance and sponsorship supporting this research.

\bibliography{custom}

\appendix
\section{Reproducibility (Frozen-Corpus Protocol)}
\label{app:reproducibility}

This appendix documents the frozen-corpus evaluation protocol and cache artifact identities.
\textbf{We do not release code in this preprint.}

\subsection{Dataset Fingerprint}
TruthfulQA (generation task), validation split, $N=817$ questions, loaded via the HuggingFace \texttt{truthful\_qa} dataset (v0.0.0).

\subsection{Frozen Mode (0 Live Web Calls)}
All benchmark runs are executed with live web search disabled. Each method retrieves only from a prebuilt frozen cache and treats it as read-only during evaluation.

\subsection{Cache Artifact Identities (Not Redistributed in This arXiv Upload)}
Our frozen caches contain third-party snippet text returned by a web-search provider at collection time.
To avoid redistributing third-party content, we do not include these cache files as arXiv ancillary data.
For auditability, we record their immutable identities (file size and SHA256) below.

\paragraph{Cache A (Baseline).}
File: \texttt{frozen\_union\_A\_nokill.jsonl}. \\
Size: 13,060,342 bytes. \\
SHA256: a0be5c65f9c69e3c848541439\-1b97eecd22861041a7c13e9ed72db1898a525a1

\paragraph{Cache B (FVA-RAG).}
File: \texttt{frozen\_union\_B\_withkill.jsonl}. \\
Size: 22,796,421 bytes. \\
SHA256: 4d4241ae5cbcafee7d9fc3f6c7\-5c60dd1765c718f9954573f7a91bbe6d55127f

\paragraph{Integrity check.}
Users can verify artifact integrity by recomputing SHA256 over each JSONL file and matching the hashes above.

\end{document}